%% file: main.tex
\documentclass[11pt,letterpaper]{article}

\usepackage[margin=1in]{geometry}

\usepackage{amsmath,amsfonts,amssymb}
\usepackage{accents}

\usepackage{graphicx}
\usepackage{array}
\usepackage{booktabs}
\usepackage{multirow}
\usepackage{tablefootnote}
\usepackage[caption=false]{subfig}

\usepackage{algorithm}
\usepackage{algorithmic}

\usepackage{textcomp}
\usepackage{url}
\usepackage{verbatim}
\usepackage{comment}

\usepackage[numbers,sort&compress]{natbib}

\usepackage{xcolor}
\usepackage{hyperref}
\hypersetup{
    colorlinks=true,
    linkcolor=black,
    citecolor=blue,
    urlcolor=blue,
    pdftitle={Free-Energy-Gated Plasticity for Real-Time Online Motor Learning in Physical Human-Robot Interaction},
    pdfauthor={Hiroki Sawada and Jun Tani}
}

\newcommand{\RNum}[1]{\uppercase\expandafter{\romannumeral #1\relax}}

\begin{document}

\title{
    \textbf{Free-Energy-Gated Plasticity for Real-Time Online Motor Learning in
    Physical Human-Robot Interaction}
}

\author{
    Hiroki Sawada\thanks{
        Cognitive Neurorobotics Research Unit,
        Okinawa Institute of Science and Technology Graduate University,
        Okinawa, Japan 904-0302.
        Email: \texttt{hiroki.sawada1@oist.jp}
    }
    \and
    Jun Tani\thanks{
        Cognitive Neurorobotics Research Unit,
        Okinawa Institute of Science and Technology Graduate University,
        Okinawa, Japan 904-0302.
        Email: \texttt{jun.tani@oist.jp}.
        Corresponding author.
    }
}

\date{}

\maketitle

\input{0_abstract.tex}


\input{1_introduction.tex}

\input{2_model.tex}

\input{3_experiment.tex}

\input{4_discussion.tex}

\input{5_conclusion.tex}

\input{supplementary.tex}

\bibliographystyle{IEEEtran}
\bibliography{references}

\end{document}

%% file: 0_abstract.tex
\begin{abstract}
Fully online embodied learning requires synaptic adaptation to acquire new behaviors while preserving previously learned dynamics during ongoing interaction.
We extend the Predictive-Coding-inspired Variational Recurrent Neural Network (PV-RNN) to continuously adapt its synaptic weights and propose Free-Energy-Gated Plasticity (FEGP), which regulates the effective learning rate according to variational free energy.
In real-time physical human-robot interaction, a randomly initialized network acquired three cyclic motor patterns without offline pretraining, replay, or task-boundary signals, with all three patterns emerging in autonomous rollouts.
Controlled experiments over ten randomized teaching streams and five network initializations per stream showed that FEGP substantially improved repertoire coverage and retention of previously acquired patterns after they left the recent observation window.
Neither a constant learning rate matched to the gate's time-averaged effective rate nor replay of the same gain values with disrupted temporal organization reproduced these improvements.
These results indicate that the temporal allocation of plasticity relative to model-environment mismatch, rather than simply its average magnitude or distribution, is critical for maintaining previously acquired behaviors during continued online learning.
\end{abstract}

%% file: 1_introduction.tex
\section{Introduction}
\label{sec:introduction}
Humans and animals acquire motor skills through continuous interaction with their environment, without a clear separation between inference and learning. 
Perception, prediction, action, and synaptic modification unfold together within an ongoing sensorimotor loop, such that adaptation directly shapes subsequent actions and observations. 
This differs from many artificial learning systems in which learning and generation are temporally separated and weight updates are performed offline or episodically. 
To distinguish the setting considered here from such approaches, we refer to it as \textit{fully online}, \textit{real-time}, and \textit{boundary-free}: weights are updated at every sensory sample without a separate training phase, each model update is completed before the next sensory sample is processed, and no task identity or transition signal is supplied to the learner.
The central challenge in such a setting is not merely rapid adaptation, but regulation of adaptation. 
Synaptic updates must remain sufficiently plastic to acquire new behavioral dynamics while limiting interference with those already learned, reflecting the classical stability-plasticity dilemma \cite{grossberg1987competitive} and its manifestation as catastrophic forgetting in neural networks. 
This challenge is particularly important for embodied agents because weight changes affect subsequent actions, which in turn alter the observations that drive future adaptation. 
Fully online embodied learning therefore requires a mechanism that regulates when and how strongly synaptic adaptation occurs.

Modern continual learning (CL) research has largely focused on mitigating catastrophic forgetting. 
Regularization-based methods such as Elastic Weight Consolidation \cite{kirkpatrick2017overcoming} and Synaptic Intelligence \cite{zenke2017continual} constrain parameter changes according to their estimated importance to previously acquired knowledge, while later approaches have relaxed assumptions about task labels or explicit task boundaries \cite{schwarz2018progress, aljundi2018memory, aljundi2019task}.
These methods progressively reduce the amount of task information required during learning, but they have not generally addressed the acquisition of recurrent motor dynamics from scratch in a fully online, real-time, and boundary-free embodied setting. 
In such a setting, the learner's own actions shape the observations that drive subsequent adaptation, coupling continual learning directly to ongoing physical interaction.

Replay-based continual-learning methods mitigate forgetting by revisiting stored examples or regenerated past experience \cite{lopez2017gradient, shin2017continual, buzzega2020dark}. 
Meta-learning provides another approach by learning initial conditions or mechanisms that support rapid or selective adaptation, as in MAML \cite{finn2017model} and ANML \cite{beaulieu2020learning}. 
These approaches demonstrate effective ways to preserve or allocate plasticity, but they rely on retained or regenerated past experience, or on offline meta-training over a predefined task distribution. 
They therefore differ from the setting considered here, where plasticity must be regulated directly from signals available during ongoing interaction, without replay or prior meta-training.

In robotics and physical human-robot interaction (pHRI), motor learning is often framed as learning from demonstration (LfD) \cite{argall2009survey}, where demonstrations are collected before policy or model optimization.
Dynamic Movement Primitives \cite{schaal2006dynamic}, for example, represent movements as stable dynamical systems whose parameters are fitted to recorded demonstrations. 
Recent continual robot learning has also explored task-structured benchmarks, skill libraries, and mechanisms for preserving previously acquired dynamics \cite{wan2024lotus, liu2023libero, auddy2026scalable}. 
These approaches differ from the setting considered here, in which state inference, motor generation, physical interaction, and synaptic adaptation proceed continuously within a single uninterrupted sensorimotor process.

A key open problem in recurrent embodied learning is therefore how to regulate synaptic plasticity so that multiple motor patterns can be acquired from scratch during uninterrupted, real-time interaction without interfering with previously acquired dynamics.

A natural way to regulate plasticity online is to use a mismatch signal that is already available within the inference process. 
Predictive coding provides such a framework by continuously comparing predictions with sensory observations \cite{rao1999predictive}, and related free-energy formulations extend this idea by expressing inference and learning in terms of variational free-energy minimization \cite{friston2010free}. These principles have also been applied to robotic control and active inference \cite{da2022active, liu2025hierarchical}. 
The Predictive Coding inspired Variational Recurrent Neural Network (PV-RNN) \cite{ahmadi2019novel} provides a concrete implementation for temporal sequences and has been applied to goal-directed behavior and physical human-robot interaction \cite{matsumoto2020goal, matsumoto2022goal, sawada2024human}. 
During online interaction, PV-RNN performs inference through error regression over a sliding window of recent observations, adapting its latent states so that the generative model better explains the observed sequence. 
Its synaptic weights, however, remain fixed after training, preventing the recurrent dynamics themselves from being learned during interaction.

Allowing synaptic weights to adapt online would enable the recurrent dynamics themselves to change as new behaviors are encountered. 
In real-time interaction, however, inference and weight updates must be completed within each model-update interval, making it impractical to optimize over the entire interaction history. 
Adaptation must therefore rely on a finite window of recent observations. 
Because each weight update is then driven primarily by this recent context, continuous adaptation can bias the model toward recent behavior and interfere with previously acquired dynamics. 
Fully online learning therefore requires not only synaptic adaptation, but also a mechanism that regulates when and how strongly that adaptation is expressed.

Biological and computational accounts of learning suggest that the rate of synaptic change can be regulated by internally generated signals of prediction error, surprise, or uncertainty. Prediction errors have been linked to adaptive updates of internal models \cite{friston2017active}, while surprise- and uncertainty-dependent regulation of plasticity has been studied through metaplastic synapses \cite{iigaya2016adaptive}, neuronal-surprise signals \cite{barry2024fast}, hierarchical Bayesian filters \cite{mathys2014uncertainty}, and neuromodulatory mechanisms that amplify learning under unexpected observations \cite{angela2005uncertainty}. 
Related work in predictive-coding robotics has further shown that internally generated prediction errors can reflect confidence in ongoing perceptual inference \cite{sawada2025cernet}. 
These accounts motivate treating plasticity as a dynamically regulated quantity driven by model-environment mismatch rather than as a fixed learning rate. 
Recent formulations of synaptic learning as free-energy minimization provide further theoretical motivation for linking plasticity regulation to variational free energy \cite{kim2024bayesian}. 
Importantly, variational free energy is already computed within PV-RNN during online inference, making it a natural internal signal for regulating synaptic adaptation without requiring an external task or novelty signal.

We therefore propose Free-Energy-Gated Plasticity (FEGP), which scales synaptic updates as a sigmoidal function of variational free energy. 
We use variational free energy as an internally generated signal of model-environment mismatch, increasing plasticity under substantial mismatch and suppressing it as the model comes to explain the recent interaction better.
FEGP thus couples synaptic adaptation directly to a signal already computed within the predictive-coding inference process.

The contributions of this work are threefold.
1) We establish and evaluate a fully online embodied learning setting without task boundaries, replay, or offline consolidation, and show that unregulated online adaptation provides little retention of previously acquired motor patterns.
2) We introduce FEGP, a sigmoidal mechanism that regulates synaptic plasticity according to variational free energy. FEGP improves repertoire coverage and retention, and controlled comparisons show that this effect depends on the temporal allocation of plasticity relative to model-environment mismatch.
3) We demonstrate the incremental acquisition of multiple motor behaviors from scratch during continuous physical human-robot interaction, with all three instructed patterns emerging in autonomous model rollouts.
\footnote{A video of a complete teaching session showing the robot learning online during physical human-robot interaction is available at \url{https://www.youtube.com/watch?v=QOgh2rTOnaw}.\label{fn:video}}

Together, these results show that internally generated mismatch signals can regulate plasticity online to support continual motor learning during ongoing physical interaction.

%% file: 2_model.tex
\section{Method}
\label{sec:method}

\subsection{Model and Fully Online Learning Formulation}

The proposed model builds on the Predictive-Coding-inspired Variational Recurrent Neural Network (PV-RNN) framework \cite{ahmadi2019novel}. 
The generative architecture and physical interaction scheme follow the formulation described in \cite{sawada2024human}. 
Here, we describe the modifications required to extend PV-RNN from online state inference to fully online synaptic learning.

Conventional PV-RNN separates synaptic learning from online interaction. 
Synaptic weights are first optimized offline using prerecorded sensory sequences. 
During subsequent interaction, the weights remain fixed, while posterior latent variables are adapted through online error regression over a sliding window of recent observations.

We remove this phase separation. The network is initialized with random weights and receives no offline pretraining. 
During interaction, posterior-state inference, synaptic adaptation, and future generation are performed repeatedly within the same online learning process.

Let $\mathbf{X}$ denote a sequence of sensory observations and $\mathbf{z}$ denote the latent variables of the generative model. 
Learning and inference are formulated by minimizing the variational free energy,
\begin{equation}
\begin{aligned}
\mathcal{F}
=
D_{\mathrm{KL}}\!\left[
q_\phi(\mathbf{z}\mid\mathbf{X})
\Vert
p_\theta(\mathbf{z})
\right]
-
\mathbb{E}_{q_\phi(\mathbf{z}\mid\mathbf{X})}
\left[
\log p_\theta(\mathbf{X}\mid\mathbf{z})
\right],
\end{aligned}
\label{eq:free_energy}
\end{equation}
where $p_\theta$ denotes the generative model parameterized by $\theta$, and $q_\phi$ denotes the variational posterior parameterized by $\phi$. 
The first term penalizes divergence between the posterior and prior latent distributions, while the second penalizes inaccurate sensory predictions.

For online inference, this objective is evaluated over a sliding window of recent observations. 
At model update $t$, the newest sensory observation is appended to the window, which contains the most recent $N$ observations. 
Once the window reaches length $N$, the oldest observation is removed as each new observation is added. 
The posterior latent variables are then optimized over the current window, allowing the internal state to adapt to the recent sensory history.

The sliding window provides the finite context required for online inference and synaptic adaptation. 
It is not an episodic replay buffer: it contains only the current recent history and does not store completed demonstrations, separate episodes, or a long-term memory of past experience.

In the fully online formulation, the synaptic parameters are also optimized during every model update. 
Conceptually, without plasticity modulation, a synaptic update can be written as
\begin{equation}
\theta_{t+1}
=
\theta_t
-
\alpha
\nabla_{\theta}
\mathcal{F}_t,
\label{eq:weight_update}
\end{equation}
where $\alpha$ denotes the base learning rate and $\mathcal{F}_t$ denotes the free-energy objective evaluated over the current past window. 
In the implementation, these updates are performed with Adam over multiple optimization iterations within each model update. 
The plasticity mechanism introduced in Sec.~\ref{sec:fegp} scales the effective optimizer step size applied to the synaptic parameters.

Posterior-state inference, synaptic adaptation, motor generation, and physical interaction therefore form a continuing closed-loop sensorimotor process. 
The model-generated motor target affects the robot's subsequent state, and the resulting sensory observation is incorporated into the next model update.

\begin{figure*}[htbp]
  \centering
  \includegraphics[width=\textwidth]{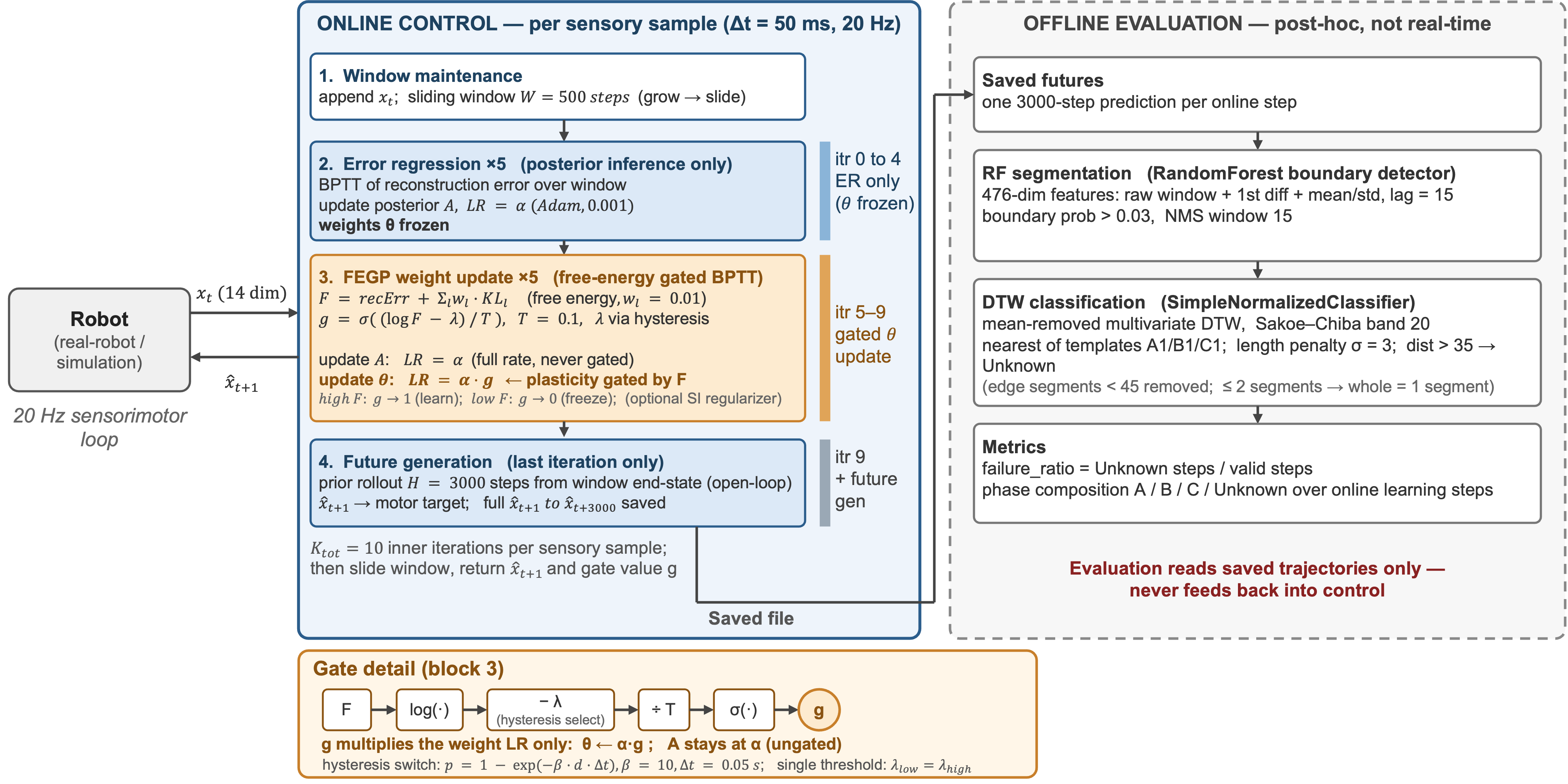}
  \caption{
  Overview of the fully online learning loop and the offline evaluation pipeline.
  Left: each model update incorporates the newest sensory observation into the sliding past window, performs posterior-state inference through error regression, adapts the synaptic weights through gated BPTT, and generates a future trajectory for motor control and subsequent analysis.
  Right: saved trajectories are segmented and classified post hoc using the evaluation pipeline described in Sec.~\ref{sec:eval_pipeline}.
  The evaluation pipeline does not feed information back into the online learning loop.
  The numerical settings used in the experiments are listed in Table~\ref{tab:hyperparams}.
  }
  \label{fig:computation_schematics}
\end{figure*}

Fig.~\ref{fig:computation_schematics} summarizes this online learning process and the separate evaluation pipeline. 
As shown on the left, each model update incorporates the newest observation into the sliding window, optimizes the posterior latent variables through error regression, adapts the synaptic parameters, and generates a future trajectory from the resulting internal state. 
As shown on the right, saved trajectories are segmented and classified offline for quantitative evaluation. 
This evaluation is strictly post hoc and provides no observations, targets, gradients, or other feedback to the online learning process.

\subsection{Naive Online Weight Adaptation}

Without plasticity modulation, Eq.~\ref{eq:weight_update} applies the same base learning rate at every model update.
Because each gradient is computed from the current sliding window, the resulting weight change is driven primarily by prediction errors in the recent interaction history.
This formulation does not distinguish between transient deviations within an already represented behavior and sustained mismatches that may require modification of the underlying dynamics.
It also provides no state-dependent mechanism for reducing plasticity after the recent observations have become predictable.
Consequently, recently observed patterns can continue to drive synaptic changes even when such changes interfere with dynamics acquired earlier in the interaction.
We refer to this constant-plasticity formulation as naive online weight adaptation.
Its behavior is evaluated experimentally against the proposed gated formulation.

To regulate when and how strongly synaptic adaptation occurs, we introduce a free-energy-dependent plasticity gain.

\subsection{Free-Energy-Gated Plasticity}
\label{sec:fegp}

We introduce Free-Energy-Gated Plasticity (FEGP), which regulates synaptic adaptation according to variational free energy. 
We use variational free energy as an internally generated signal of model-environment mismatch: plasticity is increased when the mismatch is large and suppressed as the recent interaction becomes better explained by the existing dynamics.

The effective learning rate applied to the synaptic parameters is defined as
\begin{equation}
\alpha_{\mathrm{eff}}(t)
=
\alpha\,g(s_t),
\label{eq:alpha_eff}
\end{equation}
where $\alpha$ is the base learning rate, $s_t$ is the free-energy-based gate signal defined below, and $g(s_t)\in(0,1)$ is the plasticity gain.

\paragraph{Gate signal}

Let $\mathcal{W}_t$ denote the set of time steps in the current sliding window, with $|\mathcal{W}_t|\le N$. 
Each of the $D=14$ sensory dimensions is represented by $U=10$ softmax units. 
The accuracy term used for the gate is the mean prediction divergence over the current window, summed over sensory dimensions and softmax units,
\begin{equation}
\mathcal{F}^{\mathrm{acc}}_t
=
\frac{1}{|\mathcal{W}_t|}
\sum_{k\in\mathcal{W}_t}
\sum_{d=1}^{D}
\sum_{u=1}^{U}
\tilde{p}_{k,d,u}
\ln
\frac{\tilde{p}_{k,d,u}}
{\hat{x}_{k,d,u}},
\label{eq:gate_acc}
\end{equation}
where $\tilde{p}_{k,d,u}$ and $\hat{x}_{k,d,u}$ denote the softmax-transformed sensory target and the corresponding model prediction, respectively.

The free-energy quantity used for plasticity regulation additionally includes the meta-prior-weighted complexity term,
\begin{equation}
\bar{\mathcal{F}}_t
=
\mathcal{F}^{\mathrm{acc}}_t
+
\sum_l
w_l\,
\overline{\mathrm{KL}}_{l,t},
\label{eq:gate_fe}
\end{equation}
where $w_l$ is the meta-prior of layer $l$ and $\overline{\mathrm{KL}}_{l,t}$ is the posterior-prior divergence of that layer averaged over the current window.

The gate signal is defined as the natural logarithm of the window-averaged variational free energy,
\begin{equation}
s_t
=
\ln\!\left(
\bar{\mathcal{F}}_t+\epsilon_F
\right),
\qquad
\epsilon_F=10^{-12},
\label{eq:gate_signal}
\end{equation}
where $\epsilon_F$ prevents taking the logarithm of zero. Averaging over the current window keeps the scale of the gate signal consistent while the window is still growing. The gate thresholds introduced below are defined on this logarithmic scale.

\paragraph{Plasticity gain}

The plasticity gain is defined as a sigmoid function of the gate signal,
\begin{equation}
g(s_t)
=
\sigma\!\left(
\frac{s_t-\lambda}{T}
\right)
=
\frac{1}{
1+\exp\!\left(
-\frac{s_t-\lambda}{T}
\right)
},
\label{eq:sigmoid_gate}
\end{equation}
where $\lambda$ is the gate-signal value at which $g(s_t)=0.5$, and $T$ controls the sharpness of the transition.

When $s_t$ exceeds $\lambda$, the effective learning rate approaches the base learning rate. 
When $s_t$ falls below $\lambda$, synaptic adaptation is progressively suppressed. 
The temperature $T$ determines whether the transition between lower and higher plasticity is sharp or gradual.

\paragraph{Application to the optimizer}

Each model update performs $K_{\mathrm{tot}}$ optimization iterations. 
The first $K_{\mathrm{ER}}$ iterations update only the posterior variables, whereas the remaining $K_W$ iterations update both the posterior variables and the synaptic parameters. 
The gate signal and plasticity gain are recomputed from the current window at every optimization iteration.

\begin{figure}[tbph]
  \centering
  \includegraphics[width=0.95\linewidth]{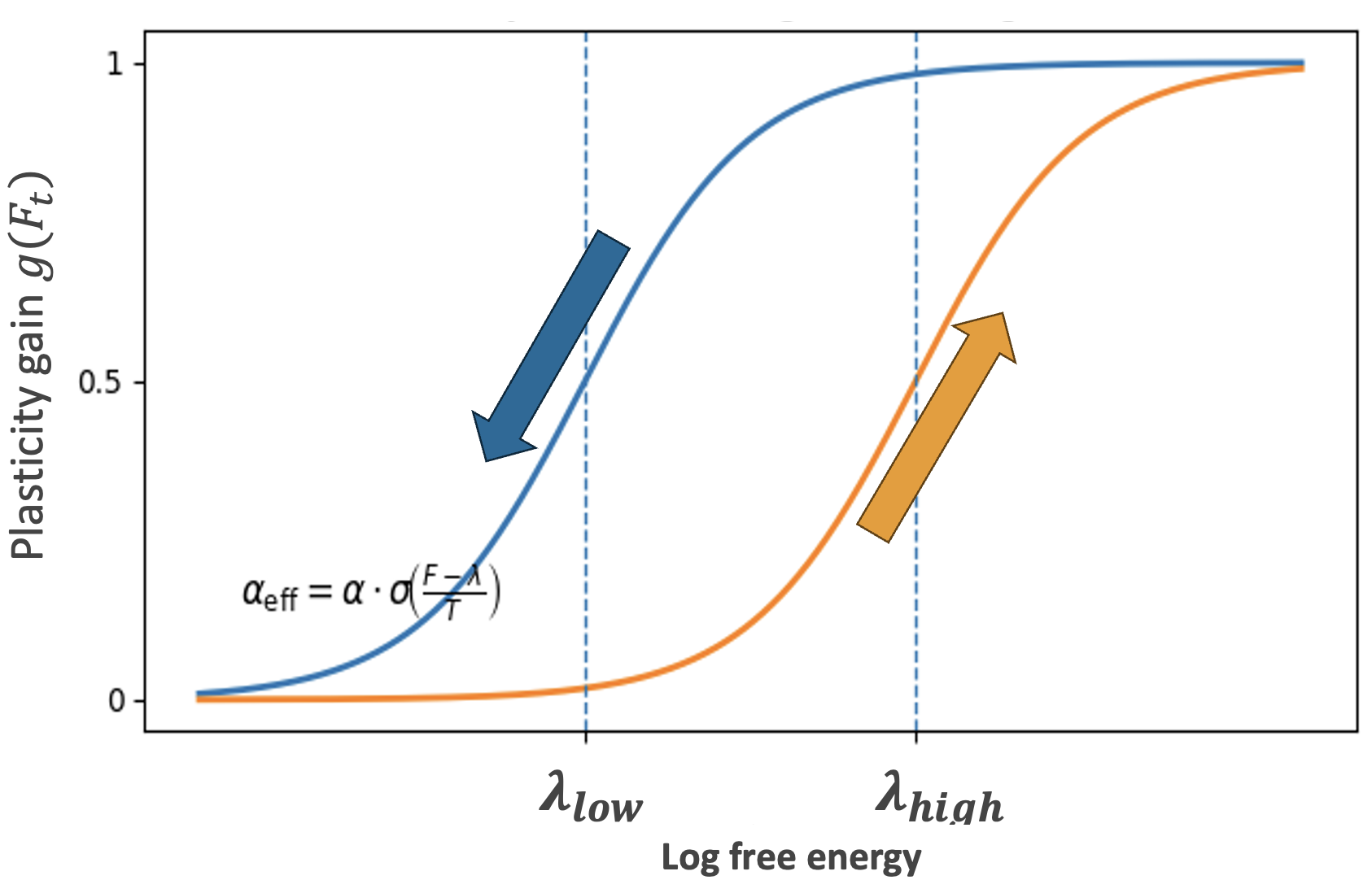}
  \caption{
  Schematic illustration of Free-Energy--Gated Plasticity.
  The plasticity gain $g(s_t)$ is a sigmoid function of the gate signal $s_t$, defined as the logarithm of the window-averaged variational free energy (Eq.~\ref{eq:gate_signal}).
  In the single-threshold formulation, one threshold $\lambda$ determines the transition between lower and higher plasticity.
  In the hysteretic formulation, the active threshold is selected from $\lambda_{\mathrm{low}}$ and $\lambda_{\mathrm{high}}$ according to the current plasticity regime.
  The single-threshold gate is recovered when $\lambda_{\mathrm{low}}=\lambda_{\mathrm{high}}=\lambda$.
  }
  \label{fig:Methods_FegpSchematics}
\end{figure}

The basic FEGP mechanism is specified by a single threshold $\lambda$ and a temperature $T$. 
It allocates plasticity selectively over time by increasing synaptic adaptation when the gate signal is high and suppressing it when the gate signal is low. 
For sufficiently small $T$, the transition becomes sharp and the gain approaches zero when $s_t$ remains below $\lambda$, making synaptic updates negligible once the recent interaction is well predicted.
However, a low-temperature single-threshold gate can switch rapidly when $s_t$ fluctuates near $\lambda$. 
We therefore introduce a hysteretic extension that adds temporal persistence by using separate thresholds for entering and leaving the adaptive regime. 
The single-threshold formulation remains the basic FEGP mechanism, while the hysteretic formulation is used to reduce rapid switching of the gain.
Fig.~\ref{fig:Methods_FegpSchematics} illustrates the relationship between the single-threshold and hysteretic formulations. 
In the single-threshold case, $\lambda_{\mathrm{low}}=\lambda_{\mathrm{high}}=\lambda$. 
In the hysteretic formulation, the active threshold depends on the current plasticity regime, as described below.

\paragraph{Hysteretic regime persistence}
\label{sec:hysteresis}

In the single-threshold formulation, the gain is determined directly by the current gate signal.
When $s_t$ repeatedly crosses $\lambda$, the gain can therefore alternate rapidly between low and high values.

The hysteretic extension introduces a binary regime state $r_t\in\{0,1\}$ that persists across optimization iterations. 
The stable regime ($r_t=0$) uses the higher threshold $\lambda_{\mathrm{high}}$, whereas the adaptive regime ($r_t=1$) uses the lower threshold $\lambda_{\mathrm{low}}$, with $\lambda_{\mathrm{low}}<\lambda_{\mathrm{high}}$. 
In the stable regime, a relatively large gate signal is therefore required to engage substantial plasticity. 
Once the adaptive regime has been entered, plasticity remains engaged until the gate signal falls below the lower threshold.

A regime transition is permitted only when the relevant threshold is crossed and the gate signal is changing in the direction associated with the target regime. 
We measure this direction using
\begin{equation}
\Delta s_t
=
s_t-\bar{s}_t,
\label{eq:gate_direction}
\end{equation}
where $\bar{s}_t$ is the mean gate signal over the most recent $N_F$ gate evaluations. 
A transition from the stable to the adaptive regime requires $s_t>\lambda_{\mathrm{high}}$ and $\Delta s_t>0$, whereas a transition from the adaptive to the stable regime requires $s_t<\lambda_{\mathrm{low}}$ and $\Delta s_t<0$.

When these conditions are satisfied, the corresponding transition is sampled once per optimization iteration according to
\begin{equation}
P_{\mathrm{stable}\rightarrow\mathrm{adaptive}}
=
1-\exp\!\left[
-\beta\,(s_t-\lambda_{\mathrm{high}})\,\Delta t
\right],
\label{eq:transition_stable_to_adaptive}
\end{equation}
and
\begin{equation}
P_{\mathrm{adaptive}\rightarrow\mathrm{stable}}
=
1-\exp\!\left[
-\beta\,(\lambda_{\mathrm{low}}-s_t)\,\Delta t
\right],
\label{eq:transition_adaptive_to_stable}
\end{equation}
where $\beta$ controls how rapidly the transition probability increases as the gate signal moves beyond the active threshold.

A stochastic transition rule allows the dwell time in each regime to vary smoothly with the extent to which the threshold is exceeded, rather than switching immediately at the first crossing. 
The regime is initialized in the adaptive state, and a fixed random seed is used for the transition draws so that repeated runs are deterministic under otherwise identical conditions.

The hysteretic formulation is intended to stabilize the temporal behavior of the plasticity gain rather than to change which behaviors can be acquired. 
Its effect on regime-switching dynamics and task-level performance is evaluated against the single-threshold formulation in the controlled experiments. 
The numerical settings for the model, online inference, and hysteretic extension are summarized in Table~\ref{tab:hyperparams}.

Algorithm~\ref{alg:online_update} summarizes one complete model update using the hysteretic formulation. 
The single-threshold formulation is obtained by using a fixed threshold $\lambda$ and omitting the regime-transition and threshold-selection steps.

\begin{algorithm}[htpb]
\caption{One model update with FEGP}
\label{alg:online_update}
\begin{algorithmic}[1]
\STATE \textbf{input:} observation $\mathbf{X}_t$, synaptic parameters $\theta$, posterior variables $\mathbf{A}$, regime state $r$
\STATE append $\mathbf{X}_t$ to the past window; drop the oldest step if the window exceeds $N$
\FOR{$i=1$ \TO $K_{\mathrm{tot}}$}
  \STATE forward pass over the window; obtain $\mathcal{F}^{\mathrm{acc}}$ and $\overline{\mathrm{KL}}_l$
  \STATE $\bar{\mathcal{F}} \leftarrow \mathcal{F}^{\mathrm{acc}} + \sum_l w_l \overline{\mathrm{KL}}_l$
  \STATE $s \leftarrow \ln(\bar{\mathcal{F}}+\epsilon_F)$
  \STATE update $\bar{s}$ over the last $N_F$ evaluations; $\Delta s \leftarrow s-\bar{s}$
  \STATE $r \leftarrow \textsc{RegimeTransition}(r,s,\Delta s,\lambda_{\mathrm{low}},\lambda_{\mathrm{high}},\beta)$
  \STATE $\lambda_{\mathrm{active}} \leftarrow \lambda_{\mathrm{high}}$ if $r$ is stable, else $\lambda_{\mathrm{low}}$
  \STATE $g \leftarrow \sigma\!\left((s-\lambda_{\mathrm{active}})/T\right)$
  \STATE backpropagate through time over the window
  \STATE $\mathbf{A} \leftarrow \mathrm{Adam}(\mathbf{A},\nabla_{\mathbf{A}}\bar{\mathcal{F}};\alpha)$ \COMMENT{ungated}
  \IF{$i>K_{\mathrm{ER}}$}
    \STATE $\theta \leftarrow \mathrm{Adam}(\theta,\nabla_{\theta}\bar{\mathcal{F}};\alpha g)$ \COMMENT{gated}
  \ENDIF
\ENDFOR
\STATE generate an $H$-step open-loop rollout from the resulting internal state
\STATE \textbf{return} motor target for step $t$, rollout
\end{algorithmic}
\end{algorithm}

\subsection{Robotic Implementation}
\label{sec:robotic_implementation}

The physical experiments were conducted using a dual-arm humanoid robot, Torobo\footnote{Torobo by Tokyo Robotics: https://robotics.tokyo/products/torobo/.}. 
The PV-RNN operates in end-effector space. 
For each arm, the model predicts a seven-dimensional pose consisting of a three-dimensional Cartesian position and a four-dimensional unit quaternion, resulting in a 14-dimensional model output.

The predicted end-effector poses are converted into joint-angle targets through inverse kinematics and executed by the compliant low-level controller described in \cite{sawada2024human}. 
The controller allows the experimenter to guide the robot kinesthetically while tracking the model-generated targets.

The measured joint angles are converted back into end-effector poses through forward kinematics. 
The resulting 14-dimensional pose vector is used as the sensory observation $\mathbf{X}_t$ supplied to the PV-RNN. 
The model therefore learns and predicts trajectories in end-effector space, whereas physical actuation is performed in joint space. 
The low-level controller is kept fixed across all experimental conditions.

%% file: 3_experiment.tex
\section{Experiments}

Two experiments were conducted to evaluate fully online synaptic adaptation and the proposed Free-Energy-Gated Plasticity (FEGP) mechanism. 
Experiment~1 uses prerecorded kinesthetic teaching sequences to analyze how FEGP regulates online learning under controlled observation streams. 
Experiment~2 evaluates the complete system during real-time physical human-robot interaction. 
In both experiments, the PV-RNN was initialized with random weights and operated without offline pretraining.

\subsection{Common Setup}
\label{sec:exp_setup}

The instructed behaviors consisted of three periodic two-arm movements, denoted A, B, and C, each beginning and ending at a home posture with both arms extended in front of the robot. 
Pattern A consisted primarily of lateral motion: the arms swung outward and returned, with each hand travelling approximately $50$~cm sideways while moving less than $4$~cm vertically.
Patterns B and C formed a mirror-image pair. 
In pattern B, the left hand rose by approximately $50$~cm while the right hand descended by approximately $50$~cm, with the directions reversed in pattern C. 
All three patterns included a forward-backward excursion of approximately $19$-$27$~cm. 
Thus, B and C were closely related mirror-image movements, whereas A was distinguished primarily by its lateral motion.

Both experiments used the sensorimotor representation described in Sec.~\ref{sec:robotic_implementation}.
Table~\ref{tab:hyperparams} summarizes the network, inference, and optimization settings shared across experimental conditions.

\begin{table}[htbp]
\centering
\caption{Shared model, inference, and optimization settings.}
\label{tab:hyperparams}
\begin{tabular}{lll}
\toprule
Component & Parameter & Value \\
\midrule
\multicolumn{3}{l}{\textit{Network}} \\
Sensory layer & softmax units, $\sigma$ & 10, 0.05 \\
PV-RNN L1 & $d$, $z$, $\tau$ & 40, 4, 3 \\
PV-RNN L2 & $d$, $z$, $\tau$ & 20, 2, 9 \\
Both PV-RNN layers & meta-prior $w$ & 0.01 \\
\midrule
\multicolumn{3}{l}{\textit{Online inference and learning}} \\
Past-window length & $N$ & 500 \\
Open-loop rollout horizon & $H$ & 3000 \\
Optimizer & Adam & -- \\
Base learning rate & $\alpha$ & $1\times10^{-3}$ \\
Adam coefficients & $\beta_1,\beta_2$ & 0.9, 0.999 \\
Adam constant & $\epsilon$ & $10^{-4}$ \\
Optimization iterations & $K_{\mathrm{tot}}$ & 10 \\
Posterior-only iterations & $K_{\mathrm{ER}}$ & 5 \\
Posterior and weight iterations & $K_{\mathrm{W}}$ & 5 \\
\bottomrule
\end{tabular}
\end{table}

\subsection{Teaching Data}
\label{sec:teaching_data}

First, a human experimenter physically guided both robot arms through the three periodic movement patterns A, B, and C, and ten individual movement cycles were recorded for each pattern. 
The duration of one cycle ranged from 46 to 58 model updates, with a median of 52.

Experiment~1 used ten teaching streams of 6{,}000 model updates constructed from these prerecorded kinesthetic demonstrations. 
Each teaching stream was assembled by concatenating cycles sampled from the recorded pool. 
After each completed cycle, the stream switched pattern with probability 0.2, and on a switch the next pattern was drawn uniformly from the two patterns not currently being taught. 
Otherwise, the current pattern was repeated using another randomly selected cycle of the same class. 
Because the successor of a pattern was not fixed, a model that had learned a single long cyclic sequence could not reproduce the stream, and the first two cycles of each stream were restricted to patterns A and B so that the third pattern was always introduced during the interaction.
Because the streams were assembled from labeled cycles, their exact cycle boundaries and class identities were known and could be used as ground truth for evaluator validation without manual annotation. 
The same ten teaching streams were supplied to every experimental condition to ensure a common observation sequence for comparison.

Experiment~2 used live kinesthetic teaching during physical interaction, as described in Sec.~\ref{sec:exp2_robot}.

\subsection{Evaluation Pipeline}
\label{sec:eval_pipeline}

All quantitative trajectory measures were computed using a fixed evaluation pipeline separate from the learning model.
Generated trajectories were first segmented into individual movement cycles and then classified as movement A, B, C, or \textit{Unknown}.
The same pipeline and fixed parameters were used for all conditions.
The relationship between this evaluation pipeline and the online learning loop is illustrated in Fig.~\ref{fig:computation_schematics}.

\paragraph{Segment-boundary detection}

Segment boundaries were detected using a 200-tree random forest trained on labeled cycle boundaries.
The input features were computed from local pose windows and included the pose values, their frame-to-frame differences, and the mean and standard deviation within each window, yielding 476 features with a window lag of 15 model updates.

Boundary probabilities were thresholded at 0.03 and then subjected to non-maximum suppression within a 15-update neighborhood.
Trajectories with no reliable internal boundary were treated as single segments, while incomplete edge segments shorter than 45 updates were excluded.

\paragraph{Segment classification}

Each detected segment was classified using multivariate dynamic time warping (DTW) with a Sakoe-Chiba band of 20 steps \cite{sakoe1978dynamic}.
Before computing DTW, the temporal mean was removed separately from the segment and the reference trajectory. No amplitude normalization was applied, and quaternion components were treated as ordinary real-valued dimensions.

For each pattern, the reference trajectory was defined as the length-normalized mean of its ten recorded demonstrations. The same demonstrations were used to estimate the cycle-length statistics of that pattern.
Each segment was assigned to the class with the minimum DTW distance and was accepted as an instance of that class only if two conditions were satisfied: its DTW distance was at most $\tau=0.90$, and its duration lay within three standard deviations of the class mean. Otherwise, the segment was labeled \textit{Unknown}.
The two criteria were applied separately because they detect different types of mismatch. DTW is tolerant to temporal warping and therefore cannot distinguish a correct spatial trajectory executed at an implausible tempo, whereas a duration criterion cannot detect an incorrect trajectory with a plausible cycle length.

The threshold $\tau$ was calibrated using the recorded demonstrations and 280 control trajectories constructed to violate different aspects of the instructed movements, as described in Sec.~S1 of the supplementary material. 
At $\tau=0.90$, none of the recorded cycles were rejected, whereas 85.7\% of the control trajectories were rejected. 
Nearly all accepted controls were time-reversed trajectories, reflecting the limited sensitivity of the present DTW-based shape measure to movement direction.

\paragraph{Evaluator validation}

Because all reported trajectory measures depend on this pipeline, we validated it against the known cycle boundaries and class labels of the teaching streams.
Boundary detection achieved $F_1=1.000$ at a tolerance of three model updates, and the classifier correctly assigned all ground-truth cycles, with an end-to-end accuracy of 0.998 using automatically detected boundaries.
To assess whether this performance depended on using the same recorded demonstrations for constructing and evaluating the pipeline, we additionally performed a demonstration-level hold-out validation. 
High boundary-detection and classification performance was maintained on streams assembled from held-out demonstrations.
Full validation results, including concatenation controls, negative-control trajectories, threshold calibration, and the hold-out analysis, are reported in Sec.~S1 of the supplementary material.

\paragraph{Metrics}

\textit{Repertoire coverage} is the fraction of online-learning steps for which all three instructed movement classes appear within the model's 3{,}000-step open-loop rollout.

\textit{Retention} is the fraction of rollouts in which a class appears while that class is absent from the observation window, pooled over all such periods. 
A class counts as absent at a given update if it occurred earlier in the stream but not within the preceding $N=500$ updates. Retention is additionally reported resolved by the elapsed duration of the absence.

\textit{Shape distance} is the median DTW distance of the accepted segments in a run to the reference of their assigned class.
Whereas repertoire coverage measures which instructed patterns are present in a rollout, shape distance measures how closely the accepted generated cycles follow the demonstrated paths. 
The two measures can therefore vary independently.

\textit{Unidentified-segment ratio}, used in Experiment~2, is the fraction of segments in an open-loop rollout labeled \textit{Unknown}.

\subsection{Experiment 1: Controlled Analysis of Free-Energy-Gated Plasticity}
\label{sec:exp1}

Experiment~1 was organized around three main objectives. 
First, we identified the operating range of the plasticity gate and selected the FEGP configuration used in the subsequent controlled comparisons and Experiment~2. 
Second, we tested the functional contribution of FEGP using nested controls that separately matched its average effective learning rate and the distribution of its gain values, allowing us to distinguish free-energy-dependent plasticity allocation from these simpler alternatives. 
Third, we examined how availability depended on how long a pattern had been absent from the observation window, and how the acquired repertoire developed over the session. 
We additionally characterized how hysteresis affected the temporal dynamics of the gate and compared FEGP with Synaptic Intelligence as a representative continual-learning method.

All acquisition conditions included online synaptic adaptation. 
At each model update, the model generated a 3{,}000-step open-loop rollout from its current inferred state, which was evaluated using the segmentation and classification pipeline described in Sec.~\ref{sec:eval_pipeline}.

Preliminary runs placed the gate signal $s_t$ approximately between $-9.5$ and $-6.5$.
The single-threshold sweep was therefore chosen to cover and extend beyond this range and was used to identify the interval of $\lambda$ over which acquisition proceeded.
Within that interval, the hysteretic gate was swept over pairs of thresholds, and the setting $(\lambda_{\mathrm{low}},\lambda_{\mathrm{high}})=(-9,-7)$ was selected.
This configuration was used for all controlled comparisons in Experiment~1 and for the physical interaction in Experiment~2, and is referred to below simply as FEGP.
Table~\ref{tab:exp1_conditions} summarizes the experimental conditions.

\begin{table}[htbp]
\centering
\caption{Conditions in Experiment~1.}
\label{tab:exp1_conditions}
\begin{tabular}{ll}
\toprule
Condition & Setting \\
\midrule
FEGP & $(\lambda_{\mathrm{low}},\lambda_{\mathrm{high}})=(-9,-7)$, $T=0.1$ \\
Constant-plasticity baseline & $g(s_t)=1$ \\
Single-threshold sweep & $\lambda\in\{-10.0,-9.5,\ldots,-6.0\}$, $T=0.1$ \\
Hysteretic-gate sweep & $\lambda_{\mathrm{low}}\in\{-10,-9,-8\}$, \\
& $\lambda_{\mathrm{high}}\in\{-8,-7,-6\}$ \\
Constant-rate control & $\alpha_{\mathrm{const}}=\overline{\alpha g(s_t)}$ under FEGP \\
Replayed-gain control & recorded $g(s_t)$ replayed in shuffled, \\
& shifted, or reversed order \\
Synaptic Intelligence & $c\in\{0,10^{-6},10^{-5},\ldots,10^{-1},1.0\}$ \\
\bottomrule
\end{tabular}
\end{table}

Each run covered a full 6{,}000-model-update teaching stream, with the generated open-loop rollout scored every five model updates. 
All controlled comparisons were evaluated across ten teaching streams and five network initializations per stream, giving 50 sessions per condition, while the threshold and hysteresis sweeps used one initialization per stream.

\textit{Threshold and hysteresis analysis.}
We first varied the single threshold $\lambda$ in increments of 0.5 to identify the range in which acquisition proceeded. 
Within this operating range, combinations of $\lambda_{\mathrm{low}}\in\{-10,-9,-8\}$ and $\lambda_{\mathrm{high}}\in\{-8,-7,-6\}$ with $\lambda_{\mathrm{low}}<\lambda_{\mathrm{high}}$ were evaluated for both task-level performance and the temporal stability of the gain.
The selected setting was used as the reference FEGP configuration for the comparisons that follow.

\textit{Time-averaged learning-rate control.}
For each stream, a constant learning rate was set equal to the time-averaged effective rate produced by FEGP on that stream, isolating temporally varying plasticity from a reduced average rate.

\textit{Replayed-gain control.}
Whereas the time-averaged learning-rate control preserves only the average effective learning rate, this control preserves the gain values themselves while altering their temporal order. 
The gain sequence produced by FEGP was replayed after uniform permutation, a half-period shift, or time reversal. 
These transformations preserve the set of gain values while disrupting their alignment with the model-environment mismatch.

\textit{Absence duration.}
On the standard streams, availability was additionally resolved by how long the pattern had been absent from the observation window, and the repertoire was resolved by model update, because both conditions were still improving at the end of the session.

\textit{Comparison with Synaptic Intelligence.}
For comparison with an existing continual-learning method, Synaptic Intelligence (SI) \cite{zenke2017continual} was evaluated under constant plasticity. 
The regularization coefficient was selected from
$c\in\{0,10^{-6},10^{-5},\ldots,10^{-1},1.0\}$,
with a decay factor of $\gamma=1-10^{-6}$ and a stabilizing constant of $\xi=10^{-6}$.
Because the present setting provides no task boundaries, the importance estimate cannot be consolidated at the end of a task.
We therefore used a continuously accumulating variant, whose update rule is given in Sec.~S2 of the supplementary material; conclusions about SI in this paper refer to that implementation.

\subsection{Experiment 2: Online Acquisition During Physical Interaction}
\label{sec:exp2_robot}

Experiment~2 evaluates the complete system during real-time physical human-robot interaction using the FEGP configuration $(\lambda_{\mathrm{low}},\lambda_{\mathrm{high}})=(-9,-7)$ selected in Experiment~1. 
Posterior inference, synaptic adaptation, and motor generation were executed in real time at 20~Hz.

A human experimenter physically guided both arms through patterns A, B, and C, advancing to the next pattern in the cyclic order with probability 0.2 after each completed cycle. 
The live sessions therefore followed the cyclic order, whereas the prerecorded streams of Experiment~1 randomized the successor; comparisons between the two experiments are made only qualitatively. 
Four independent 5-min sessions, corresponding to 6{,}000 model updates each, were conducted under the same protocol.
At each model update, the model generated a 3{,}000-step open-loop rollout from its current inferred state. 
The unidentified-segment ratio and shape distance defined in Sec.~\ref{sec:eval_pipeline} were computed from these rollouts. 
A recording of one such session, with the rollout segmentation and classification displayed alongside the interaction, is provided in the supplementary video introduced in Sec.~\ref{sec:introduction} (footnote~\ref{fn:video}).

To examine the development of autonomous generative dynamics independently of ongoing sensory input, model parameters were saved at 100-update intervals. 
At each checkpoint, the saved parameters were fixed and the model generated 1{,}000 steps autonomously without external sensory observations. 
These trajectories were segmented and classified using the same evaluation pipeline. 
The checkpoints at 1{,}200, 2{,}000, and 6{,}000 model updates are shown as representative examples, while the class-count and occupancy summaries use all checkpoints.

%% file: 4_discussion.tex
\section{Results and Discussion}

The results support four main conclusions. First, FEGP enabled the acquisition of three motor patterns from scratch during real-time physical human-robot interaction. 
Second, the gate did not change what the model produced while a pattern was being taught; it changed how long that pattern remained available after teaching moved on. 
Third, this effect arose from the temporal allocation of plasticity relative to model-environment mismatch, and not from a reduction in the average effective learning rate or from the distribution of gain values alone. 
Fourth, the breadth of the generated repertoire developed late in the session and was still increasing when teaching ended, so it is reported as a trajectory rather than as a single late average.

\subsection{Experiment 1: Controlled Analysis of FEGP}
\subsubsection{Operating Regime and Hysteretic Gate Dynamics}
\label{sec:results_lambda}

The single-threshold sweep revealed a bounded operating range in which online acquisition proceeded successfully. 
Acquisition was maintained over $\lambda\in[-10,-7.5]$, with all-three-pattern coverage ranging from 14.1-31.7\% during the late interaction, but collapsed at $\lambda\geq-6.5$, where 99.2\% of the generated segments could no longer be matched to the instructed patterns and coverage fell to 0.3\%. 
Within the working range, both coverage and retention were highest at $\lambda=-8$ (31.7\% and 10.5\%). 
Lower thresholds left the gate open too readily, so plasticity was barely regulated, whereas higher thresholds suppressed it before the repertoire had been acquired. Detailed results of the single-threshold sweep are shown in Figs.~S1 and S2.

\begin{figure*}[tbph]
  \centering
  \includegraphics[width=0.95\textwidth]{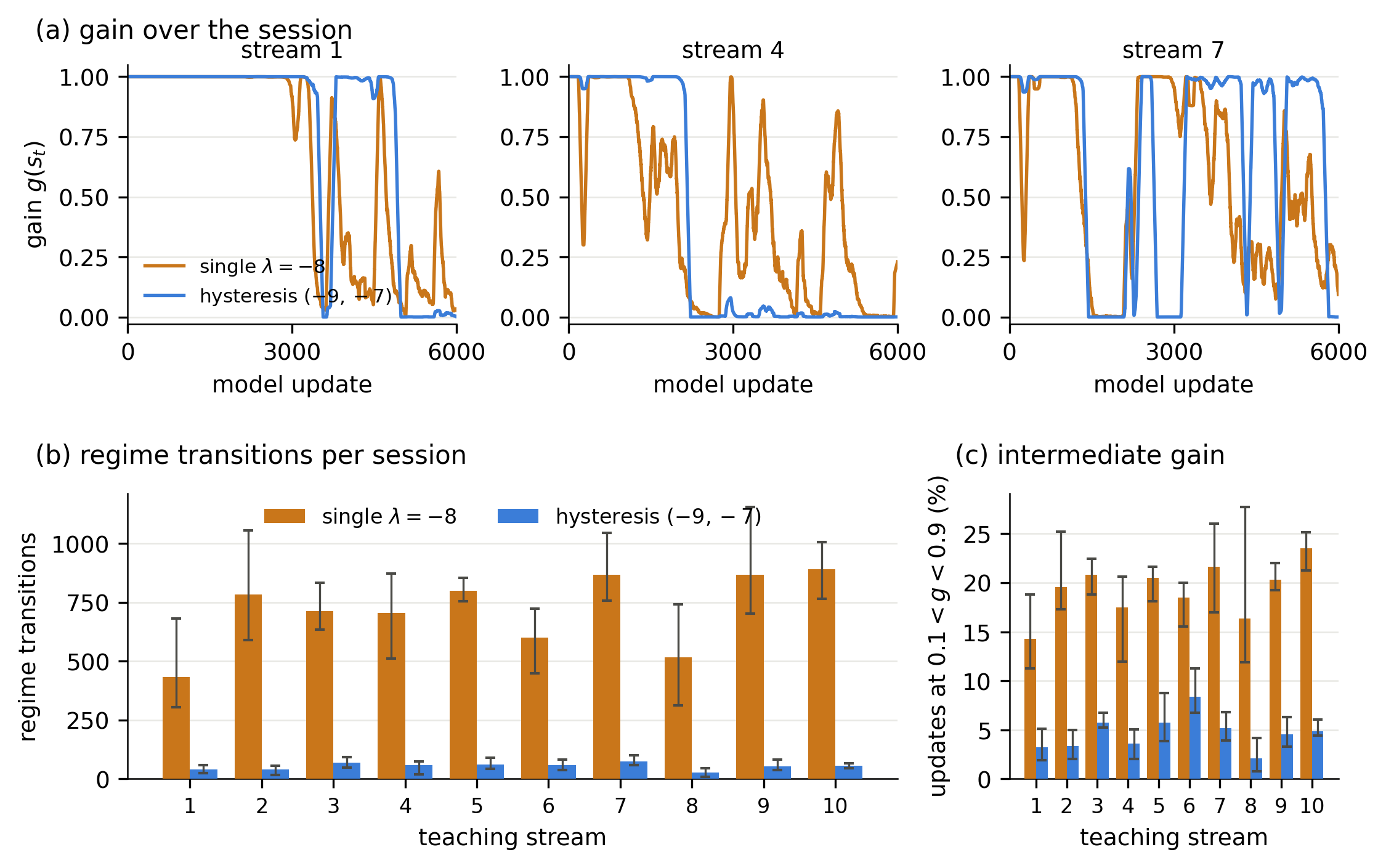}
  \caption{
  Effect of hysteresis on the temporal dynamics of the plasticity gate.
  (a) Representative gain trajectories under the single-threshold setting $\lambda=-8$ and the hysteretic setting $(\lambda_{\mathrm{low}},\lambda_{\mathrm{high}})=(-9,-7)$.
  (b) Number of threshold crossings or regime transitions during the interaction.
  (c) Fraction of model updates with intermediate gain, defined by $0.1<g(s_t)<0.9$.
  The hysteretic formulation strongly reduces rapid switching and yields more clearly separated low- and high-plasticity regimes.
  }
  \label{fig:gate_dynamics}
\end{figure*}

Although $\lambda=-8$ provided a suitable operating point, the single-threshold formulation switched frequently between lower- and higher-gain regions as the free-energy signal fluctuated around the threshold. 
Across the 50 sessions, the threshold was crossed 718 times on average (range 303-1{,}156), and 19.3\% of model updates occurred at intermediate gain values. 
Introducing separate opening and closing thresholds strongly reduced this switching: the hysteretic formulation produced 54 regime transitions on average (range 8-101) and reduced the fraction of intermediate-gain updates to 4.7\%, yielding more clearly separated low- and high-plasticity regimes (Fig.~\ref{fig:gate_dynamics}).

We next swept pairs of $(\lambda_{\mathrm{low}},\lambda_{\mathrm{high}})$ within the operating range. Pairs whose lower threshold was $-9$ or $-8$ retained absent patterns (10.6-38.9\%), whereas pairs opening at $-10$ did not (0.3-6.8\%), and pairs closing at $-6$ lost the repertoire entirely (0.0\% coverage, 84.3-94.7\% unidentified segments). 
Within the retaining band the pairs were comparable: $(-9,-7)$, the configuration deployed in Experiment~2, reached 45.9\% coverage and 23.1\% retention, and $(-8,-7)$ reached 26.3\% and 19.4\%. 
The choice within this band is therefore not critical, and $(-9,-7)$ is used as the reference FEGP configuration throughout, since it is the setting that ran on the robot. Unless otherwise stated, FEGP refers to this configuration in the remainder of the paper.

\subsubsection{What Gating Contributes: Nested Controls}
\label{sec:results_matched_lr}

We next used the controls of Experiment~1 to determine which properties of the plasticity schedule account for the effect of FEGP. Table~\ref{tab:gating_controls} summarizes repertoire coverage and retention for the principal conditions.

\begin{table}[tbph]
\centering
\caption{Repertoire coverage and retention under FEGP and its control conditions. Differences are relative to FEGP and are reported as coverage / retention in percentage points.}
\label{tab:gating_controls}
\begin{tabular}{lccc}
\toprule
Condition & Cov. (\%) & Ret. (\%) & $\Delta$ vs.\ FEGP (pp) \\
\midrule
\textbf{FEGP} & \textbf{45.9} & \textbf{23.1} & \textbf{---} \\
Constant & 18.9 & 0.0 & $-27.0 / -23.1$ \\
Matched rate & 12.5 & 0.2 & $-33.4 / -22.9$ \\
Permuted gain & 21.3 & 0.5 & $-24.6 / -22.6$ \\
Reversed gain & 11.4 & 0.8 & $-34.5 / -22.3$ \\
Shifted gain & 17.1 & 2.2 & $-28.8 / -20.9$ \\
\bottomrule
\end{tabular}
\end{table}

FEGP substantially outperformed constant plasticity, increasing repertoire coverage from 18.9\% to 45.9\% and retention from 0.0\% to 23.1\%. 
The difference was especially pronounced for retention, indicating that unregulated online adaptation provided little preservation of previously acquired patterns.

Matching the average amount of plasticity was not sufficient to reproduce this effect. The matched-rate control, which used a constant learning rate equal to the time-averaged effective rate produced by FEGP, achieved only 12.5\% coverage and 0.2\% retention. 
Thus, the improvement under FEGP cannot be explained simply by a reduction in the average effective learning rate.

Preserving the gain values while disrupting their temporal organization was likewise insufficient. 
Uniform permutation of the recorded gains reduced coverage to 21.3\% and retention to 0.5\%. 
Time reversal yielded 11.4\% coverage and 0.8\% retention, while a half-period shift yielded 17.1\% coverage and 2.2\% retention. 
All transformed schedules therefore remained substantially below FEGP despite being derived from the same recorded gain values.

Together, these controls show that the effect of FEGP depends on the temporal allocation of plasticity relative to model-environment mismatch, rather than simply on its average magnitude or on the distribution of gain values.

\subsubsection{Shape Distance of Generated Cycles}
\label{sec:results_shape}

FEGP did not improve the shape accuracy of individual generated cycles. 
The median shape distance of accepted cycles was higher under FEGP than under constant plasticity (0.475 versus 0.393), and the matched-rate control fell between them (0.459).
In physical units, the accepted cycles of both conditions followed the demonstrated paths closely: the DTW-aligned hand-position error was 40~mm under FEGP and 36~mm under constant plasticity, with orientation errors of $3.3^\circ$ and $3.0^\circ$, against 23.9~mm and $2.13^\circ$ between two human demonstrations of the same pattern. 
The 4~mm difference between the conditions is small relative to the variability of the demonstrations themselves, and it was of the same size for each of the three patterns.

These results indicate a trade-off between the accuracy of individual generated movements and the maintenance of the learned repertoire.
Constant plasticity reproduced the demonstrated trajectories more closely, whereas FEGP maintained a broader repertoire and retained patterns that were no longer present in the recent observation window (Secs.~\ref{sec:results_matched_lr} and \ref{sec:results_retention}).

\subsubsection{Retention of Patterns Absent from the Observation Window}
\label{sec:results_retention}

\begin{figure}[tbph]
  \centering
  \includegraphics[width=0.95\columnwidth]{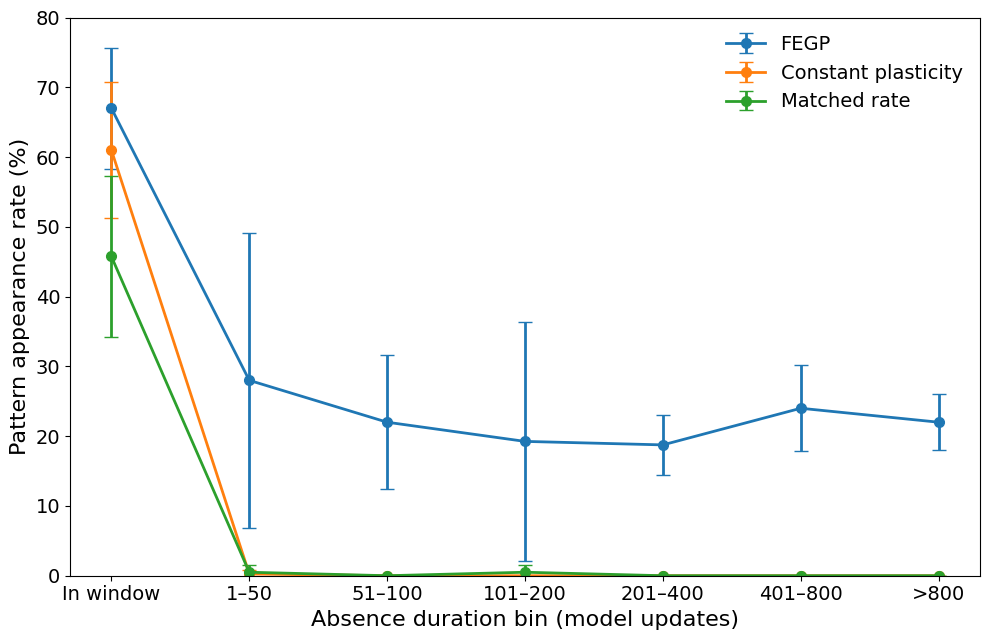}
  \caption{
  Retention of movement patterns as a function of the time elapsed since they left the 500-update observation window.
  The value at 0 corresponds to patterns still present within the observation window.
  Under FEGP, availability drops after a pattern leaves the window but remains substantially above the two constant-rate controls over longer absence intervals.
  In contrast, retention under constant plasticity and the matched-rate control falls to near zero shortly after the pattern leaves the window.
  }
  \label{fig:retention_absence}
\end{figure}

Because the teaching stream presents one movement pattern at a time, previously learned patterns regularly leave the 500-update observation window while learning continues. 
We therefore measured how often each pattern remained present in the model's rollout as a function of the time elapsed since it left the observation window (Fig.~\ref{fig:retention_absence}).

While a pattern was still present within the observation window, FEGP produced it in 66.6\% of rollouts, compared with 60.6\% under constant plasticity and 46.2\% under the matched-rate control. 
The conditions separated much more strongly after the pattern left the window. Under FEGP, retention was 27.9\% within the first 50 updates of absence and remained at 23.8\% after 51-100 updates, 23.6\% after 101-200, 24.2\% after 201-400, 32.7\% after 401-800, and 30.9\% beyond 800 updates.

Both constant-rate controls, in contrast, lost the pattern almost immediately after it left the observation window. 
Under constant plasticity, retention fell to 0.3\% within the first 50 updates of absence and was 0.0\% thereafter. 
The matched-rate control showed the same rapid loss, with 0.3\% retention within the first 50 updates, 0.0\% after 51-100 updates, 0.8\% after 101-200 updates, and 0.0\% at all longer absence intervals.

These results show that the effect of FEGP is expressed primarily after a pattern is no longer present in the recent observation history. 
Rather than disappearing shortly after leaving the observation window, previously learned patterns remained available in roughly one quarter to one third of rollouts even after several hundred updates of absence. 
FEGP therefore substantially extends the period over which previously learned movements remain available without continued rehearsal.

\subsubsection{Comparison with Synaptic Intelligence}
\label{sec:results_si}

Synaptic Intelligence showed a narrow range in which acquisition remained possible, but it did not reproduce the retention benefit of FEGP. 
For $c\leq10^{-4}$, repertoire coverage ranged from 16.8\% to 28.8\%, while retention remained at or below 0.4\%. 
The highest coverage was obtained at $c=10^{-5}$, with 28.8\% coverage, 0.4\% retention, and a 23.7\% unidentified-segment ratio. At $c=10^{-3}$, repertoire coverage fell to 0.0\% and the unidentified-segment ratio increased to 92.0\%; for $c\geq10^{-2}$, all generated segments were unidentified.

Thus, increasing SI regularization did not produce a useful trade-off between acquisition and retention in the present continuously interleaved setting. 
Weak regularization could preserve acquisition at a level comparable to the unregularized condition, but retention remained near zero, whereas stronger regularization prevented stable acquisition altogether. 
In particular, the value $c=10^{-5}$ gave the best repertoire coverage among the tested SI settings but only 0.4\% retention when used on its own.

This comparison used the continuously accumulating SI variant described in Sec.~S2 of the supplementary material. 
Because the sweep was evaluated with a single network initialization, these results are intended as a comparison with a representative continual-learning regularizer rather than as a comprehensive evaluation of SI. 
Different formulations that rely on explicit task boundaries or consolidation events may behave differently.

\subsection{Experiment 2: Real-Time Physical Human-Robot Interaction}
\label{sec:results_robot}

\begin{figure*}[tbph]
  \centering
  \includegraphics[width=0.95\textwidth]{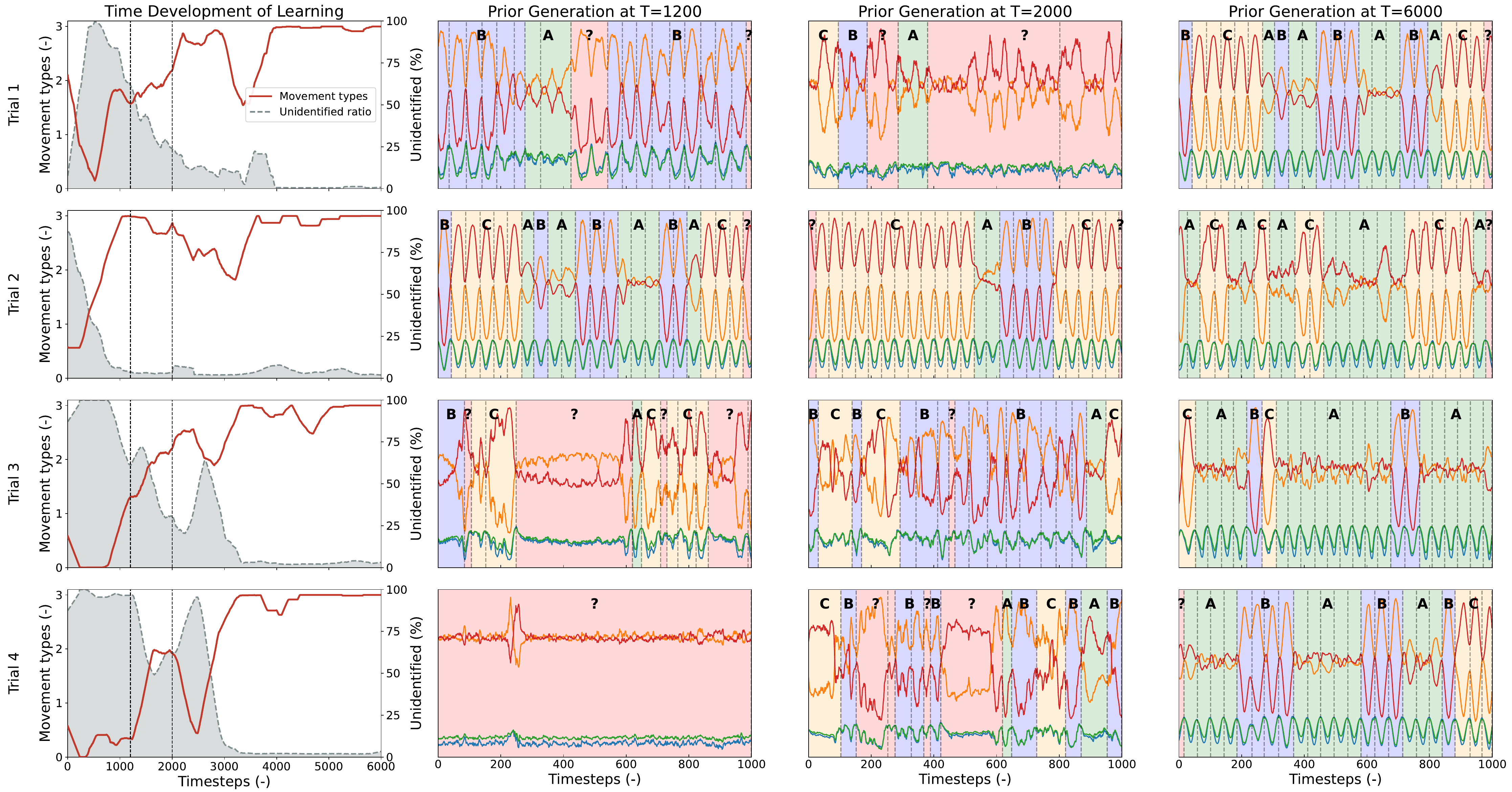}
  \caption{
  Development of the generated movement repertoire during real-time physical interaction in Experiment~2.
  Each row shows one independent session.
  Left: temporal evolution of the number of identified movement types and the unidentified-segment ratio in the open-loop rollouts generated during learning.
  Right: autonomous trajectories generated using parameters saved at model updates $t=1{,}200$, $2{,}000$, and $6{,}000$.
  The trajectories were segmented and classified using the evaluation pipeline; segments assigned to patterns A, B, and C are color-coded, while unclassifiable segments are labeled ``?''.
  }
  \label{fig:robot_learning}
\end{figure*}

Experiment~2 provides the primary embodied demonstration, using the hysteretic configuration $(\lambda_{\mathrm{low}},\lambda_{\mathrm{high}})=(-9,-7)$ during real-time physical interaction. Fig.~\ref{fig:robot_learning} shows the development of the generated repertoire across the four sessions. 
During learning, the number of identified movement types increased toward three in every session, while the unidentified-segment ratio decreased from high values early in learning to near zero toward the end of the interaction.

The autonomous trajectories generated from saved checkpoints showed the same progression. 
At $t=1{,}200$ and $t=2{,}000$, the generated trajectories frequently contained segments that could not be identified as one of the instructed movements. 
By $t=6{,}000$, these unidentified segments had largely disappeared, and all three instructed movement patterns were represented in the autonomous generation across the four sessions (Fig.~\ref{fig:robot_learning}).

The recent observation window did not contain all three patterns at every model update. 
Across the four sessions, at least one pattern was absent from the past observation window for an average of 42.5\% of the interaction, yet all three patterns became available in autonomous generation as learning progressed (Fig.~\ref{fig:robot_learning}). 
This is qualitatively consistent with the retention measured on the prerecorded streams (Sec.~\ref{sec:results_retention}), which randomized the order in which patterns were taught, whereas the live sessions followed the cyclic order.

A representative session is provided in the supplementary video (footnote~\ref{fn:video}). 
Experiment~2 therefore demonstrates the online acquisition of multiple motor behaviors from scratch during uninterrupted real-time physical human-robot interaction.

%% file: 5_conclusion.tex
\section{Conclusion}

This study demonstrated fully online acquisition of multiple motor behaviors during continuous physical human-robot interaction.
Starting from random weights and without offline pretraining or externally supplied task boundaries, the PV-RNN progressively acquired all three instructed movement patterns during four independent 5-min interaction sessions. 
As learning proceeded, the number of identified movement types increased and unidentified generated segments largely disappeared, with all three instructed patterns eventually appearing in autonomous generation.

The controlled experiments showed that Free-Energy-Gated Plasticity (FEGP) substantially improved both repertoire coverage and the retention of previously acquired motor patterns. 
This effect cannot be explained simply by reducing the average amount of plasticity. 
A constant learning rate matched to the time-averaged effective rate of FEGP produced substantially lower coverage and almost no retention. 
Likewise, replaying the same gain values after permutation, time reversal, or temporal shifting failed to reproduce the performance of FEGP. 
These controls indicate that the effect of FEGP depends on when plasticity is expressed relative to ongoing model-environment mismatch, rather than simply on the average magnitude or distribution of the plasticity gain.

The strongest effect of gating appeared after a learned pattern was no longer present in the recent observation history. 
Under constant plasticity and the matched-rate control, previously acquired patterns disappeared almost immediately after leaving the observation window. 
Under FEGP, by contrast, they remained available in a substantial fraction of model rollouts even after several hundred model updates of absence. 
FEGP therefore acts primarily by extending the availability of previously acquired behaviors while online learning continues.

This improvement did not reflect a general increase in the accuracy of individual generated movements. 
Constant plasticity reproduced the demonstrated trajectories somewhat more closely, whereas FEGP maintained a broader behavioral repertoire and substantially greater retention. 
The breadth of this repertoire also continued to increase late in the teaching session, indicating that the final repertoire was still developing when interaction ended. 
Hysteresis further stabilized the plasticity mechanism by reducing rapid switching between low- and high-plasticity regimes.

For comparison with an existing continual-learning method, we also evaluated a continuously accumulating variant of 
Synaptic Intelligence. Within the tested range, weak SI regularization allowed acquisition but provided little retention, whereas stronger regularization prevented stable acquisition. 
This comparison was based on a single network initialization and should therefore be interpreted as a representative baseline rather than as a comprehensive evaluation of Synaptic Intelligence.

Several limitations remain. The experiments considered only three periodic movement patterns, and larger repertoires, richer sensorimotor behaviors, and substantially longer interactions remain to be tested. 
The plasticity thresholds and several evaluation parameters were selected empirically and may require adaptation across tasks and embodiments. 
In addition, the present DTW-based evaluation has limited sensitivity to movement direction.

FEGP should therefore be viewed as a computational mechanism for regulating when synaptic adaptation occurs rather than as a complete solution to long-term memory or a biologically faithful model of synaptic plasticity. 
The present results nevertheless show that an internally generated model-environment mismatch signal can regulate plasticity online in a way that preserves previously acquired behavior while allowing continued learning during ongoing physical interaction. 
Future work should investigate longer-term consolidation, automatic adaptation of the gating parameters, and scaling to larger and more diverse embodied learning problems.

%% file: supplementary.tex
\clearpage

\section*{Supplementary Material}

\setcounter{section}{0}
\setcounter{equation}{0}
\setcounter{table}{0}
\setcounter{figure}{0}

\renewcommand{\thesection}{S\arabic{section}}
\renewcommand{\theequation}{S\arabic{equation}}
\renewcommand{\thetable}{S\arabic{table}}
\renewcommand{\thefigure}{S\arabic{figure}}

\section{Validation of the Evaluation Pipeline}
\label{sec:supp_eval}

The ten teaching streams of Experiment~1 were assembled by concatenating recorded movement cycles (Sec.~III-B of the main text), so their cycle boundaries and class labels are known exactly. The evaluation pipeline was validated on four representative teaching streams, comprising 459 ground-truth boundaries. 
This section reports that validation using the same fixed parameters used for all reported measurements: a random-forest boundary probability threshold of $0.03$, a non-maximum suppression window of 15 model updates, a window lag of 15, a shape threshold of $\tau=0.90$, and a duration tolerance of three standard deviations.

\subsection{Boundary detection}

Table~\ref{tab:supp_boundary} reports precision, recall, and $F_1$ over the 459 ground-truth boundaries of the four streams, as a function of the matching tolerance and of the detection threshold.

\begin{table}[htbp]
\centering
\caption{Boundary detection against ground truth (four streams, 459 boundaries).}
\label{tab:supp_boundary}
\begin{tabular}{lccc}
\toprule
Setting & Precision & Recall & $F_1$ \\
\midrule
\multicolumn{4}{l}{\textit{Matching tolerance, at threshold 0.03}} \\
$\pm3$ model updates & 1.000 & 1.000 & 1.000 \\
$\pm5$ / $\pm10$ / $\pm15$ & 1.000 & 1.000 & 1.000 \\
\midrule
\multicolumn{4}{l}{\textit{Detection threshold, at tolerance $\pm5$}} \\
0.03 (operating point) & 1.000 & 1.000 & 1.000 \\
0.10 & 1.000 & 1.000 & 1.000 \\
0.30 & 1.000 & 0.904 & 0.950 \\
0.50 & 1.000 & 0.850 & 0.919 \\
0.70 & 1.000 & 0.702 & 0.825 \\
\bottomrule
\end{tabular}
\end{table}

Because the streams are concatenations of recorded cycles, each ground-truth boundary carries a small kinematic discontinuity, and a detector could in principle exploit it.
Table~\ref{tab:supp_robust} shows that it does not: removing the discontinuity by a cosine crossfade across every join, or smoothing the whole stream so that the step change at a join becomes \emph{smaller} than the typical within-cycle step, leaves detection unchanged, as does time-warping every cycle by $\pm10\%$.

\begin{table}[htbp]
\centering
\caption{Boundary detection after removing the concatenation artifact.}
\label{tab:supp_robust}
\begin{tabular}{lccc}
\toprule
Stream & \shortstack{Boundary-step /\\ within-cycle median} & $F_1$ ($\pm5$) & Class accuracy \\
\midrule
Unmodified & 1.94 & 1.000 & 1.000 \\
Crossfade (width 9) & 1.21 & 1.000 & 1.000 \\
Smoothed (width 5) & 0.44 & 1.000 & 1.000 \\
Time-warped $\pm10\%$ & 1.96 & 1.000 & 1.000 \\
Warped $+$ crossfade & 1.25 & 1.000 & 1.000 \\
\bottomrule
\end{tabular}
\end{table}

\subsection{Segment classification and the \textit{Unknown} decision}

Given ground-truth boundaries, all 459 cycles were assigned to the correct pattern and none was rejected. End-to-end accuracy on detector-produced segments was 0.998.

Table~\ref{tab:supp_negatives} evaluates the classifier on seven categories of control trajectories that are not valid instances of the instructed patterns, with 40 trajectories per category. 
We compare the rejection rate of the rule used in this paper with that of a summed DTW-and-duration score thresholded at 35. 
Recorded cycles were evaluated with their own contribution removed from the corresponding class reference.

The comparison also illustrates why the shape and duration conditions were applied separately. 
Under the summed score, the duration term dominates the decision, causing several spatially invalid trajectories to be accepted.

\begin{table*}[!t]
\centering
\caption{Rejection performance on recorded and control trajectories.}
\label{tab:supp_negatives}
\begin{tabular*}{\textwidth}{@{\extracolsep{\fill}}lccc@{}}
\toprule
Trajectory set & Summed score, thr.\ 35 & Rule used & Shape distance (median) \\
\midrule
Recorded cycles (30) & 0.00 & 0.00 & 0.36 \\
\midrule
Time-reversed & 0.00 & 0.02 & 0.64 \\
Block-permuted & 0.00 & 0.98 & 2.35 \\
Two patterns spliced & 0.00 & 1.00 & 3.43 \\
Half cycle & 0.80 & 1.00 & 2.62 \\
Two cycles concatenated & 0.35 & 1.00 & 2.59 \\
Smoothed random walk & 0.00 & 1.00 & 2.31 \\
Static pose & 0.00 & 1.00 & 2.21 \\
\bottomrule
\end{tabular*}
\end{table*}

The 30 positives span 0.211--0.852 with a median of 0.359 and none carries a length penalty, while the 280 negatives begin at 0.303 with a median of 2.362; the area under the ROC curve of the shape term is 0.988.
The choice of $\tau$ is not delicate: rejection of the negatives varies by four percentage points over $\tau\in[0.70,1.20]$, at 89.6\%, 87.9\%, 85.7\%, 85.4\% and 85.0\% for $\tau=0.70$, $0.80$, $0.90$, $1.00$ and $1.20$, and $\tau=0.90$ is the largest value at which no recorded cycle is rejected (3.3\% are at $0.80$ and below).

The one category the rule cannot reject is time reversal, at 2\%: DTW on the present mean-removed pose representation has limited sensitivity to the direction of travel, so time-reversed movements can remain close to the same reference. 
Shape distance should therefore be interpreted primarily as agreement in path rather than in movement direction.
On these controls, the duration condition is redundant with the shape condition because every control of implausible duration is already rejected on shape. 
It is nevertheless retained because DTW is tolerant to temporal warping and could otherwise accept a correct path executed at an implausible tempo.

\subsection{Demonstration-level hold-out}

The detector, the references, the cycle-length statistics and the teaching streams all derive from the same 30 recorded cycles, and the detector's own train/validation split is a random split over sliding windows of one stream, so windows from the same cycle appear on both sides.
Table~\ref{tab:supp_holdout} repeats the validation with a split at the level of demonstrations: a training stream of 120 cycles assembled from demonstrations 1--5 of each pattern was used to fit the detector and to build the references and length statistics, and a disjoint test stream of 120 cycles assembled from demonstrations 6--10 was used for evaluation.

\begin{table}[htbp]
\centering
\caption{Evaluation on a stream assembled only from held-out demonstrations.}
\label{tab:supp_holdout}
\begin{tabular}{lcccc}
\toprule
\multicolumn{5}{l}{\textit{Boundary detection}} \\
Detector & Tol. & Prec. & Rec. & $F_1$ \\
\midrule
All 30 cycles & $\pm3$ & 1.000 & 1.000 & 1.000 \\
All 30 cycles & $\pm5$ & 1.000 & 1.000 & 1.000 \\
Demos 1--5 & $\pm3$ & 0.967 & 0.967 & 0.967 \\
Demos 1--5 & $\pm5$ & 1.000 & 1.000 & 1.000 \\
\midrule
\multicolumn{5}{l}{\textit{Cycle classification}} \\
Reference & \multicolumn{2}{c}{Thr.} & Acc. & Unknown \\
\midrule
Mean, all 30 & \multicolumn{2}{c}{0.90} & 1.000 & 0.000 \\
Mean, demos 1--5 & \multicolumn{2}{c}{0.90} & 1.000 & 0.000 \\
Single, demos 1--5 & \multicolumn{2}{c}{0.90} & 0.967 & 0.033 \\
\bottomrule
\end{tabular}
\end{table}

On the held-out stream, shape distances ranged from 0.27--0.49 against the mean-of-five reference and from 0.50--0.73 against a single held-out cycle. 
Both ranges remained below the rejection threshold $\tau=0.90$, indicating that the threshold transferred to demonstrations that took no part in building the references.



\section{Synaptic Intelligence Update Rule}
\label{sec:supp_si}

For comparison with an existing continual-learning method, we used a continuously accumulating variant of Synaptic Intelligence (SI), because the boundary-free setting provides no explicit point at which parameter importance can be consolidated.

At every synaptic update,
\begin{equation}
A \leftarrow \gamma A + (-\nabla_\theta \mathcal{F}) \odot (-\Delta\theta),
\qquad
B \leftarrow B + \Delta\theta^{\odot 2},
\label{eq:si_accum}
\end{equation}
where $\odot$ denotes elementwise multiplication. The importance estimate is recomputed after each update as
\begin{equation}
\Omega = A \oslash (B+\xi),
\end{equation}
where $\oslash$ denotes elementwise division.

A single reference parameter vector $\theta^{*}$ is fixed at initialization, and the penalty
\begin{equation}
c\sum_j \Omega_j(\theta_j-\theta^{*}_j)^2
\end{equation}
is added to the objective. We used $\gamma=1-10^{-6}$ and $\xi=10^{-6}$ in all SI conditions.


\section{Additional Figures}
\label{sec:supp_figs}

Unless otherwise noted, the multi-stream panels in this section show the same four representative teaching streams (Seq1--4), selected from the ten streams used in Experiment~1 for visual comparison across figures.

\begin{figure*}[tbph]
  \centering
  \includegraphics[width=0.8\textwidth]{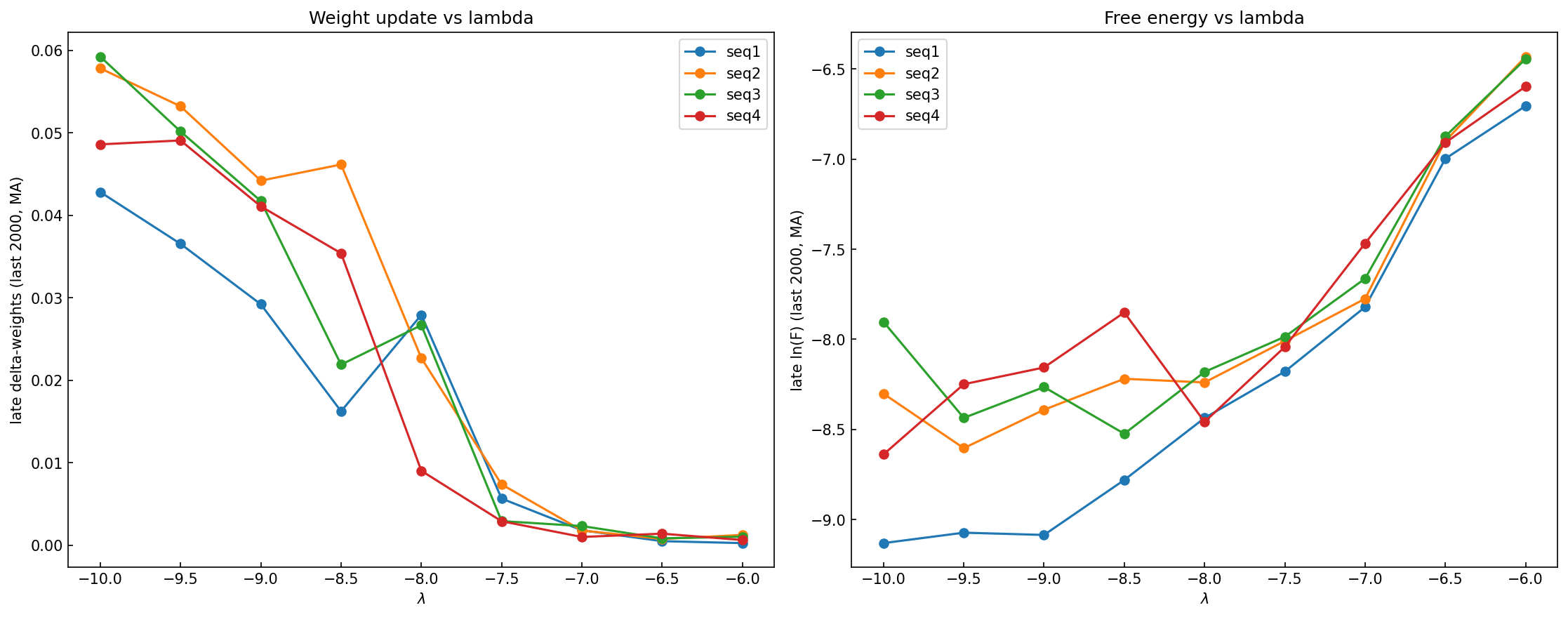}
  \caption{
    Late-stage summary of the single-threshold sweep in Experiment~1.
    Left: weight-update magnitude averaged over the final 2{,}000 model updates as a function of $\lambda$ for four representative teaching streams (Seq1--4).
    Right: corresponding late-stage log-scaled free energy for the same streams.
    Increasing $\lambda$ progressively suppresses weight adaptation while residual free energy increases.
  }
  \label{fig:supp_lambda_summary}
\end{figure*}

\begin{figure*}[tbph]
  \centering
  \includegraphics[width=0.95\textwidth]{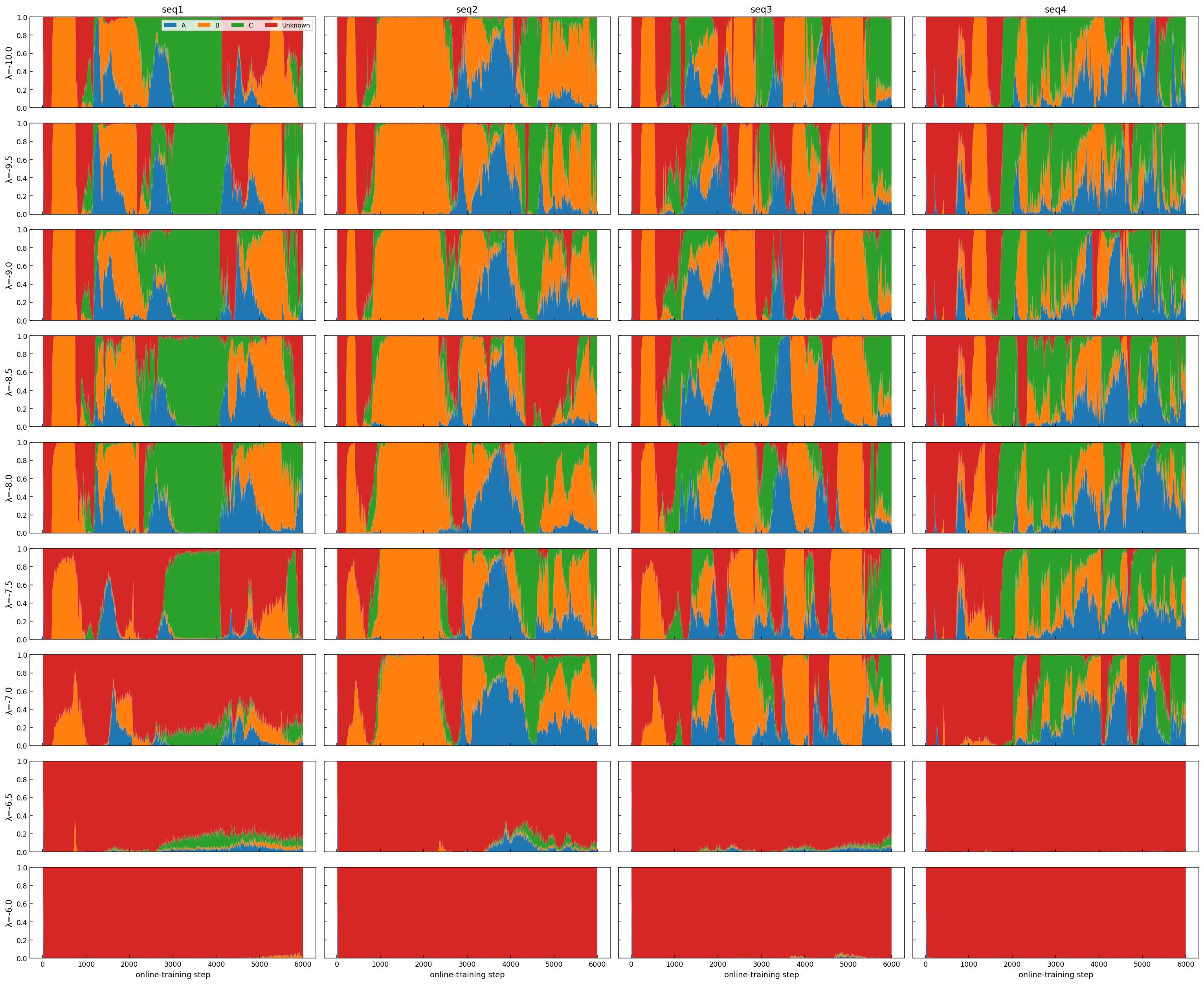}
  \caption{
  Composition of the open-loop rollout over online learning under the single-threshold plasticity gate in Experiment~1.
  Rows correspond to threshold values $\lambda\in\{-10.0,\ldots,-6.0\}$, and columns show the same four representative teaching streams (Seq1--4) used throughout this section.
  At each online-learning update, the model's 3{,}000-step rollout was segmented and classified as movement pattern A, B, C, or \textit{Unknown} using the evaluation pipeline in Sec.~III-C of the main text.
  The displayed values use a 10-update moving average.
  For $\lambda\leq-8$, the rollout develops and maintains all three instructed patterns, whereas plasticity is strongly suppressed for $\lambda\geq-6.5$.
  }
  \label{fig:lambda_phase}
\end{figure*}

\begin{figure*}[tbph]
  \centering
  \includegraphics[width=0.95\textwidth]{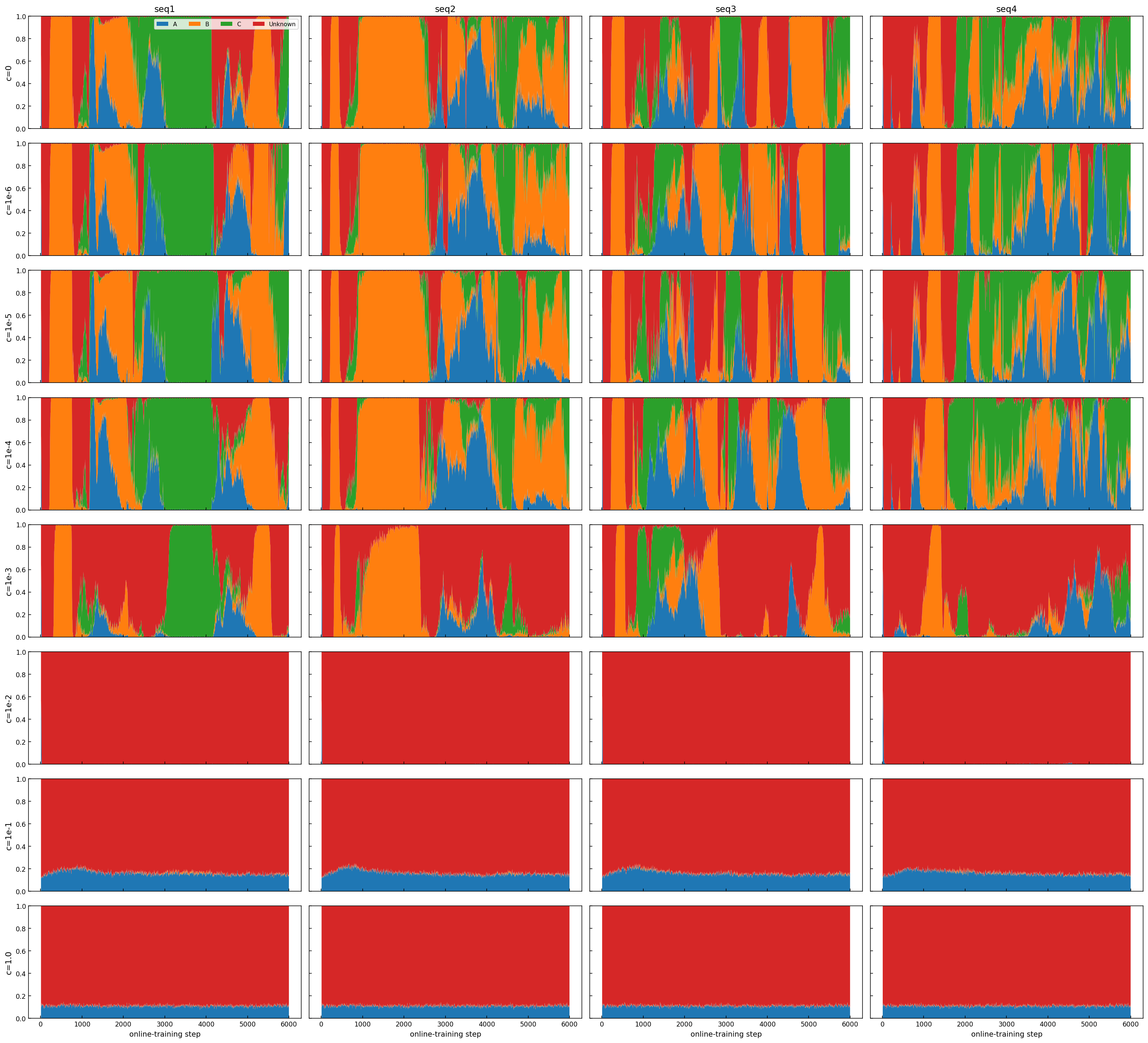}
  \caption{
  Composition of the open-loop rollout under Synaptic Intelligence with constant plasticity in Experiment~1.
  Rows correspond to Synaptic Intelligence coefficients $c\in\{0,10^{-6},10^{-5},10^{-4},10^{-3},10^{-2},10^{-1},1.0\}$, and columns show the same four representative teaching streams (Seq1--4) used in Figs.~\ref{fig:supp_lambda_summary} and \ref{fig:lambda_phase}.
  Colors are as in Fig.~\ref{fig:lambda_phase}.
  Weak regularization produces outcomes similar to the unregularized baseline, whereas strong regularization prevents stable acquisition.
  }
  \label{fig:si_sweep}
\end{figure*}